\documentclass{article}
\usepackage{spconf,amsmath,graphicx}
\usepackage{color}
\usepackage{multirow}
\usepackage{graphicx}
\usepackage{amsmath,amssymb} 
\usepackage{booktabs}
\usepackage{url}

\title{Super-Resolution information enhancement for Crowd Counting}
\name{\begin{tabular}{c}Jiahao Xie$ ^{1}$, Wei Xu$ ^{1}$, Dingkang Liang$ ^{2}$, Zhanyu Ma$ ^{1}$, Kongming Liang$ ^{1,\dagger}$, \\ Weidong Liu$ ^{3}$, Rui Wang$ ^{3}$, and Ling Jin$ ^{3}$ \end{tabular} \thanks{$ ^{\dagger}  $Corresponding author \protect\\
\indent\indent This work was supported in part by Beijing Natural Science Foundation Project No. Z200002, and in part by National Natural Science Foundation of China (NSFC) No. 62106022, 62225601, U19B2036, and in part by Beijing University of Posts, and in part by Program for Youth Innovative Research Team of BUPT, and Telecommunications-China Mobile Research Institute Joint Innovation Center.}}
\address{$ ^{1} $Beijing University of Posts and Telecommunications\\
         $ ^{2} $Huazhong University of Science and Technology\\
         $ ^{3} $China Mobile Research Institute\\
         \{xiejiahao, xuwei2020, mazhanyu, liangkongming\}@bupt.edu.cn, dkliang@hust.edu.cn,\\ \{liuweidong, wangruiyjy, jinling\}@chinamobile.com}

\begin{document}
%
\maketitle
\begin{abstract}
Crowd counting is a challenging task due to the heavy occlusions, scales, and density variations. Existing methods handle these challenges effectively while ignoring low-resolution (LR) circumstances. The LR circumstances weaken the counting performance deeply for two crucial reasons: 1) limited detail information; 2) overlapping head regions accumulate in density maps and result in extreme ground-truth values. An intuitive solution is to employ super-resolution (SR) pre-processes for the input LR images. However, it complicates the inference steps and thus limits application potentials when requiring real-time. We propose a more elegant method termed Multi-Scale Super-Resolution Module (MSSRM). It guides the network to estimate the lost details and enhances the detailed information in the feature space. Noteworthy that the MSSRM is plug-in plug-out and deals with the LR problems with no inference cost. As the proposed method requires SR labels, we further propose a Super-Resolution Crowd Counting dataset (SR-Crowd). Extensive experiments on three datasets demonstrate the superiority of our method. The code will be available at {\color{blue}\url{https://github.com/PRIS-CV/MSSRM.git}}.

\end{abstract}
\begin{keywords}
crowd counting, super-resolution, multi-scale super-resolution module, plug-in plug-out
\end{keywords}

\vspace{-10pt}
\section{Introduction}
\label{sec:intro}

Crowd counting is an essential task for understanding crowd scenes, which aims to estimate the number of persons involved in given images or videos. Traditional crowd counting algorithms~\cite{chen2012feature,ryan2009crowd,ge2009marked, chang2021your, chang2021making} achieve this task mainly by detecting individuals or regressing density-related features.

\begin{figure}[t]
\centering
\includegraphics[scale=0.07]{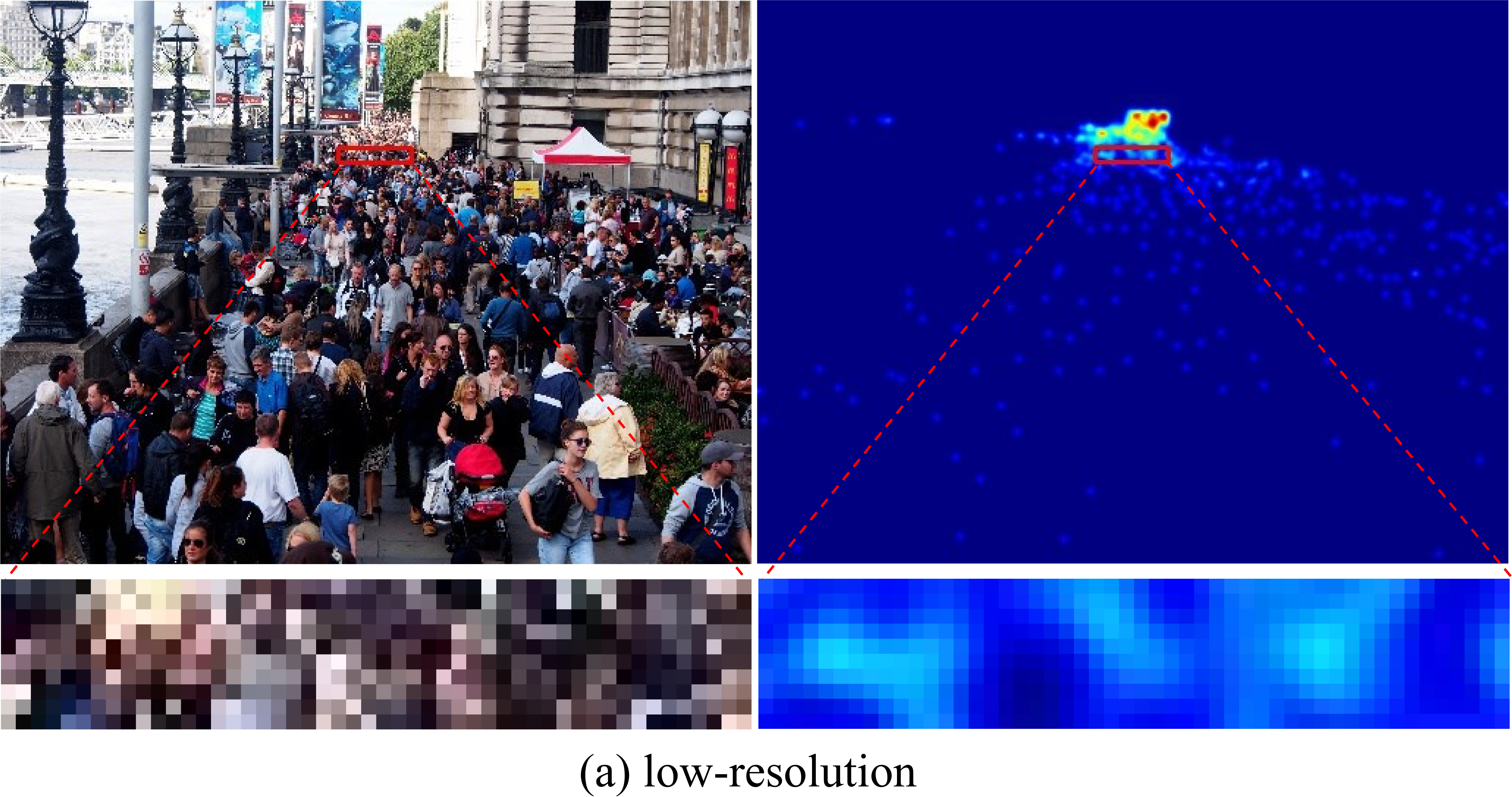}
\includegraphics[scale=0.07]{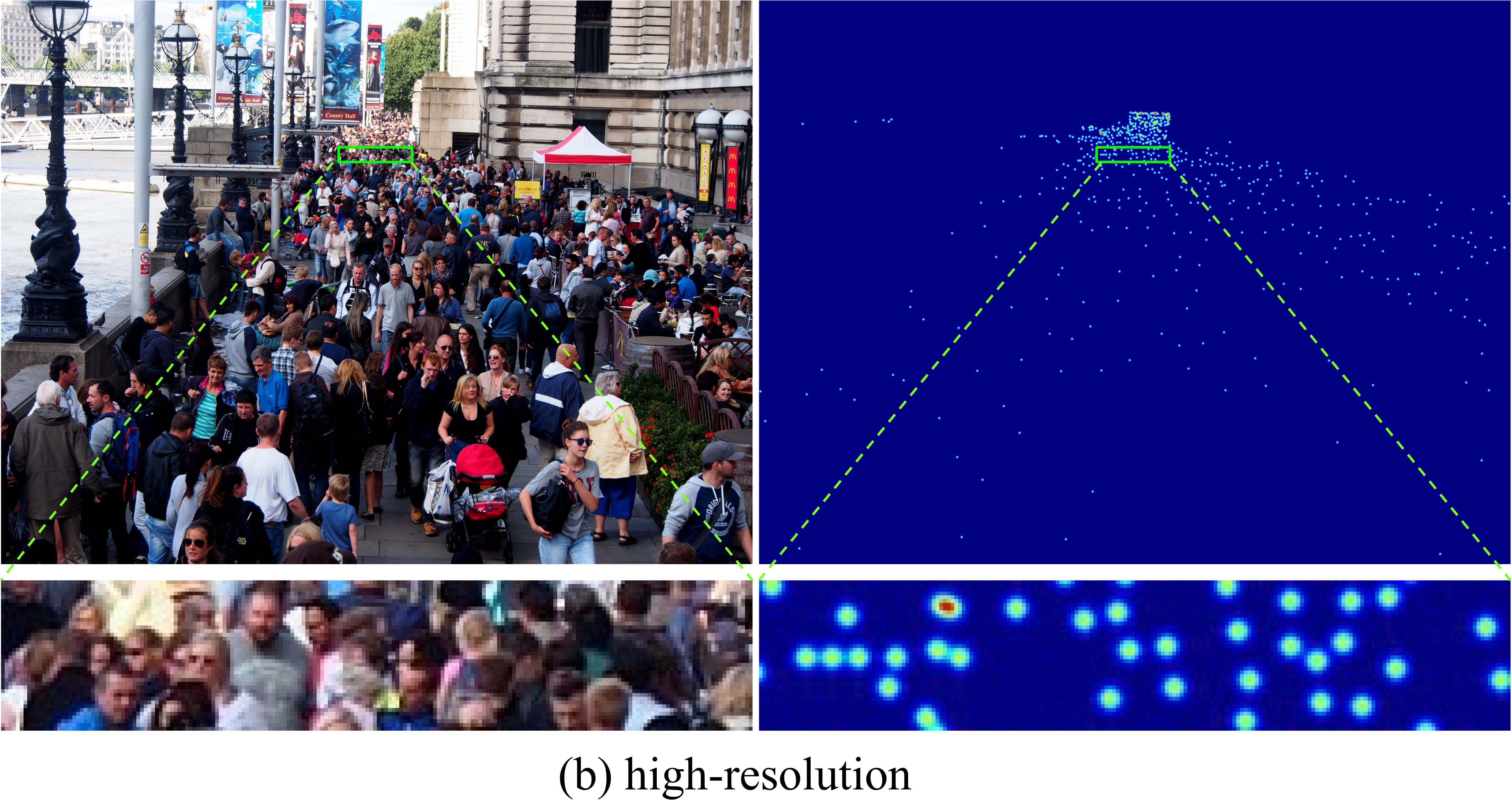}
\caption{
Comparisons of LR and HR images and corresponding density maps. (a) Persons located at dense regions of the LR image are indistinguishable. The density map is blurry. (b) The HR image contains rich detailed information. The density map is clear and sharp.
}
\vspace{-12pt}
\label{fig:intro}
\end{figure}

\begin{figure*}[t]
\centering
\resizebox{0.8\textwidth}{!}{
    \includegraphics{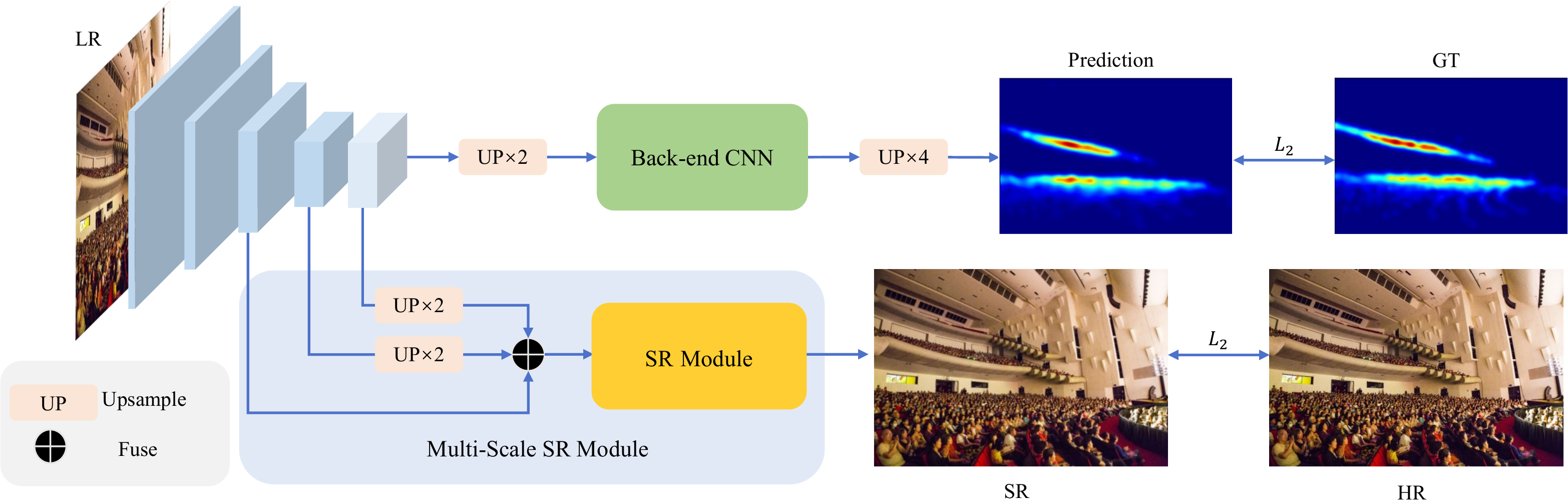}
    }
\caption{Overview of the proposed MSSRGN. Concretely, one LR image is fed into the backbone. Then, feed the output feature into two branches simultaneously, \textit{i.e.}, the back-end CNN and the MSSRM. Specifically, the back-end network generates a density map to estimate the crowd count. The MSSRM module predicts an SR image and achieves the SR task.}
\centering
\vspace{-5pt}
\label{fig:pipeline}
\end{figure*}

We have witnessed the rise of convolutional neural networks (CNN). Current mainstream counting methods~\cite{sindagi2017generating, li2018csrnet,xu2022autoscale,ranjan2018iterative,cao2018scale} adopt the CNN to regress a density map. The predicted count equals the integral of the density map, which can perform better than the traditional methods.
However, crowd counting is still a challenging task due to the large-scale variation, complex background, and blur. 
Recent methods~\cite{xu2022autoscale,jiang2020attention,cao2018scale, liang2022end, liang2022focal, liang2022transcrowd} usually focus on coping with the scale variation and background noise while ignoring the blur. Generally, low-resolution (LR) images usually present blurry phenomena, i.e., losing the massive content details of the crowd. Actually, in practical applications, LR images are really common, \textit{e.g.}, aging image capture devices, limited image storage space, downsampling to achieve real-time inference, perspective effect, occasional jitters, and so on. Networks drawback in the LR scenes are hard to extract precise semantic information when being deployed in the mentioned situations, leading to disappointing application potentials. Specifically, as shown in Fig.~\ref{fig:intro} (a), the LR image loses content details, resulting in a low-quality density map. Instead, the HR image maintains the detailed information, and the resulting density map is clear and sharp. The challenge in generating sharp density maps from LR images is the lack of detailed information.

Super-resolution~\cite{hu2019meta,wang2019textsr,lee2020srf} aims to reconstruct a visually natural HR image from its degraded LR image and has achieved great success in these years. It is a plausible method to tackle the mentioned problem.
However, employing the SR methods in the pre-processes to obtain HR images introduces extra inference costs and limits the application potentials in occasions requiring real-time estimation.

In this paper, we propose to employ the SR task as the auxiliary task to enhance the detailed information in the feature space. Specifically, we propose a Multi-Scale Super-Resolution Module (MSSRM), which enables the networks to generate HR density maps. The MSSRM works in the training phase, and guides the network to estimate reliable details even when given LR images. It could be removed during the inference phase, compensating for the loss of detail caused by blurry images without changing the original network structure. We further combine the MSSRM with mainstream backbones to propose Multi-Scale Super-Resolution Guided Network (MSSRGN). We measure the counting performance of the MSSRGN in LR circumstances to evaluate the effectiveness of our methods.

Notably, SR tasks require corresponding HR and LR labels, but existing crowd datasets capture crowd images at a single resolution level. To address this issue, we propose a new Super-Resolution Crowd dataset (SR-Crowd). 
In summary, the contributions of this paper are as follows:

1. We propose an MSSRM, which guides the network to estimate reliable detail information and boosts the counting performance in LR circumstances with no inference costs.

2. We introduce SR-Crowd, a large-scale super-resolution crowd dataset. To our knowledge, this is the first super-resolution dataset in the crowd counting area.

3. Extensive experiments on the SR-Crowd and two mainstream crowd counting datasets demonstrate the effectiveness of our method.

\vspace{-20pt}
\section{Methods}
We propose to employ the SR task as an auxiliary task to enhance the SR information for crowd counting. In this section, we introduce the SR-Crowd dataset and the MSSRGN.

\vspace{-10pt}
\subsection{Data Collection and Analysis}
The SR-Crowd images are collected from two challenging crowd datasets: NWPU-Crowd~\cite{wang2020nwpu} and UCF-QNRF~\cite{idrees2018composition}. We first extract images of height or width greater than 2048 from these two datasets. Then we adjust the corresponding ratio to make the height or width equal to 2048 as 4$\times$ upscaling images. Subsequently, we downsample the images by 2$\times$ and 4$\times$, respectively. Downsampling source images to construct a new SR dataset is a standard way in SR tasks.
Then we generate the ground-truth density maps for LR images by resizing the head center coordinates according to scaling factors. The 2$\times$ and 4$\times$ upscaling images are used as SR ground truth.

Each set of SR-Crowd data consists of 3,542 image pairs with $2,000,527$ annotations. It consists of $2,909$ images used for training and $633$ images for testing. In the dataset, the minimum and maximum counts are 0 and $17,726$, respectively, and the mean counts are $564$. The average resolution of LR images, 2$\times$ upscaling images, and 4$\times$ upscaling images are $350$ $\times$ $504$, $670$ $\times$ $1008$, and $1397$ $\times$ $2016$, respectively.
\vspace{-5pt}
\subsection{MSSRGN} We focus on improving counting performance in the LR circumstances compared to the previous work~\cite{jiang2019crowd}. As shown in Fig.~\ref{fig:pipeline}, we achieve the SR task in the training phase. Following the usual practice \cite{li2018csrnet,Ma_2019_ICCV, xu2021dilated}, we choose the pre-trained VGG-16~\cite{simonyan2014very} as the backbone. We remove the last pooling layer of the original VGG-16 to preserve spatial information and remove the full connection layer to adapt the network structure to arbitrary resolution inputs. The output size of VGG-16 is 1/8 of the original input size. Then feed the output feature to the back-end network to generate the density map. The back-end structure consists of a 3$\times$3 convolution layer and a 1$\times$1 convolution layer. 
Furthermore, we feed the features to the MSSR module to estimate the SR image and enhance the detailed information. We introduce the details of the MSSR module next.

\begin{table}[t]
\begin{center}
\small
\setlength{\tabcolsep}{3.6mm}
\resizebox{0.47\textwidth}{!}{
\begin{tabular}{lcccc}
\toprule
{\multirow{2}{*}{Interpolation method}} & \multicolumn{2} {c} {Baseline} & \multicolumn{2} {c} {MSSRGN} \\
  \cmidrule{2-5}
 & MAE$\downarrow$ & RMSE$\downarrow$ & MAE$\downarrow$ & RMSE$\downarrow$   \\
\midrule
INTER-CUBIC &159.481 &795.770 &151.908 &806.734 \\
INTER-LANCZOS4 &\underline{155.746} &\underline{765.958} &\underline{151.804} &\underline{757.212} \\
INTER-LINEAR &\textbf{153.522} &\textbf{729.571} &\textbf{148.171} &\textbf{683.561} \\
\bottomrule
\end{tabular}}
\caption{Experimental results of different resize interpolation on the SR-Crowd dataset. The first and second places are highlighted in \textbf{bold} and \underline{underline}.}
\label{tab:interpolation}
\vspace{-15pt}
\end{center}
\end{table}

\textbf{MSSRM.} We design a multi-scale super-resolution module to extract detailed information from LR images and assist the backbone network to generate high-quality density maps. MCNN~\cite{zhang2016single} proposes the multi-column convolution method to solve the multi-scale problem in the crowd. Although this method can alleviate the scale problem to a certain extent, it has a complex structure and low efficiency. Meanwhile, the number of columns limits the scale diversity of features. In this paper, we concatenate multi-scale features from the backbone as the input of the MSSRM. The super-resolution module consists of a 3 $\times$ 3 convolution layer and a sub-pixel convolution layer~\cite{shi2016real}. The 3 $\times$ 3 convolution layer reshapes the input feature to be $r^{2}$ channels where r represents upscaling factor. Then, the sub-pixel convolution rearranges the elements of tensors of $H$ $\times$ $W$ $\times$ $C$ $\times$ $r^{2}$ shape into tensors of $rH$ $\times$ $rW$ $\times$ $C$ shape according to periodic shuffling, where H, W, C denote the height, width and channels of the LR image respectively.  This method can fill in the LR features and map them to HR features. The mathematical expression can be present as:
\begin{equation}
I^{SR}=f^{n}\left(\mathbf{I}^{L R} ; W_{n}, b_{n}\right)=\xi\left(W_{n} * f^{n-1}\left(\mathbf{I}^{L R}\right)+b_{n}\right),
\nonumber
\end{equation}
where $W_n$, $b_n$, $n$ represent network weights, bias, and the number of layers n, respectively. $\xi$ is a periodic shuffling operator that rearranges elements with LR features.

\begin{table}[t]
\begin{center}
\small
\setlength{\tabcolsep}{11mm}
\resizebox{0.47\textwidth}{!}{
\begin{tabular}{lcc}
\toprule
Stage fusion & MAE$\downarrow$ & RMSE$\downarrow$   \\
\midrule
Stage 5 & 152.508 & 749.711\\
Stage 3, 5 & 156.931 & 790.493 \\
Stage 4, 5 & \underline{151.670} & \underline{739.041}\\
Stage 3, 4, 5 & \textbf{148.171} & \textbf{683.561}\\
\bottomrule
\end{tabular}}
\caption{Ablation studies on feature fusing. The first and second places are highlighted in \textbf{bold} and \underline{underline}.}
\label{tab:stage}
\vspace{-15pt}
\end{center}
\end{table}

The MSSRM contains shallow layers, which leads to limited decoding capabilities. However, it is not designed as an ideal head structure to achieve desirable SR performance on purpose. We propose the MSSRM to guide the network to enhance the detailed information in the feature space. Specifically, we supervise the SR image estimation. Transmit ground-truth lost details information to the backbone structure. Shallow layers shorten the SR information flow. It forces the encoded feature from the backbone structure to contain rich detailed information. In this way, while facing the LR circumstances, the estimated lost detail feature boosts the sharper density maps estimation performance. Note that the MSSRM will be removed after the crowd counter is well-trained. It improves the counting performance with no inference cost. Instead of employing the SR pre-processes before crowd counting, the proposed MSSRM is more elegant.


\begin{table*}
\begin{center}
\small
\setlength{\tabcolsep}{6mm}
\resizebox{0.8\textwidth}{!}{
\begin{tabular}{c|l|cc|cc|cc}
\toprule
&{\multirow{2}{*}{Method}} &\multicolumn{2}{c|}{SR-Crowd} &\multicolumn{2}{c|}{Shanghai A} &\multicolumn{2}{c}{UCF-QNRF}  \\
\cmidrule(r){3-8}
&& MAE$\downarrow$ & RMSE$\downarrow$ & MAE$\downarrow$ & RMSE$\downarrow$ & MAE$\downarrow$ & RMSE$\downarrow$ \\
\midrule
Baseline&CSRNET \cite{li2018csrnet} &157.417 & 729.247 & 90.113 & 140.453 &126.948  & \underline{213.069}\\
Ours&CSRNET\dag &{150.133} & \underline{658.396} &89.105 & 142.893 &124.445 &220.137\\
Ours&CSRNET\dag\dag &{148.712} & \textbf{616.518} &89.034 & 139.199 &121.185 &221.730\\
\midrule
Baseline&FPN \cite{xu2022autoscale}& 160.076 & 715.116 &91.393 & 140.272 &125.434 &227.319\\
Ours&FPN\dag & 153.269 & 715.735 &87.211 & \textbf{136.346} &123.165 &217.634\\
Ours&FPN\dag\dag & 158.445 & 754.565 &87.598 & \underline{137.026} &123.952 &228.763\\
\midrule
Baseline&VGG16 &153.522 &729.571 &93.351 & 141.182 &118.770 &216.013\\
Ours&MSSRGN\dag &\textbf{148.171} &683.561 &\underline{86.845} &138.164 & \textbf{115.391} & \textbf{212.518}\\
Ours&MSSRGN\dag\dag &\underline{148.564} & 696.520 & \textbf{86.664} &137.885 &\underline{115.674} &213.901\\
\bottomrule
\end{tabular}}
\caption{Comparisons of our MSSRGN with two methods on the SR-Crowd dataset, where MSSRGN\dag~and MSSRGN\dag\dag~indicate that the upscale factor are 2$\times$ and 4$\times$, respectively. The first and second places are highlighted in \textbf{bold} and \underline{underline}.}
\label{tab:comparisons}
\vspace{-8pt}
\end{center}
\end{table*}

\textbf{Loss Function.}
We use the L2 loss to supervise the density map and HR image estimations. Specifically, given the $i$-th image $X_{i}$, the loss $L(\Theta)$ can be presented as:
\begin{equation}
L(\Theta)=\frac{1}{2 N} \sum_{i=1}^{N}\left\|P\left(X_{i} ; \Theta\right)-P_{i}^{G T}\right\|_{2}^{2},
\nonumber
\end{equation}
where $\Theta$ is a learnable set of parameters. $P\left(X_{i} ; \Theta\right)$ refers to the estimated density map generated by MSSRGN. $P_{i}^{G T}$ is the ground-truth density map of image $X_{i}$. $N$ is the number of training images. $L(\Theta)$ denotes the loss between the ground-truth density map and the estimated density map.
\begin{equation}
L(\epsilon)=\frac{1}{2 N} \sum_{i=1}^{N}\left\|f\left(I_{i}^{LR} ; \epsilon\right)-I_{i}^{HR}\right\|_{2}^{2},
\nonumber
\end{equation}
where $I_{i}^{HR}$ and $I_{i}^{LR}$ refer to the HR and LR images, respectively. $\epsilon$ is a learnable set of parameters. $f\left(I^{LR} ; \epsilon\right)$ is the SR feature generated by MSSRGN. $L(\epsilon)$ denotes the loss between the HR feature and SR feature. The final objective function is defined as:
$L = L(\Theta) + \alpha L(\epsilon)$,
where $\alpha$ is a balance weight.

\begin{figure}[t]
\centering
\resizebox{0.5\textwidth}{!}{
    \includegraphics{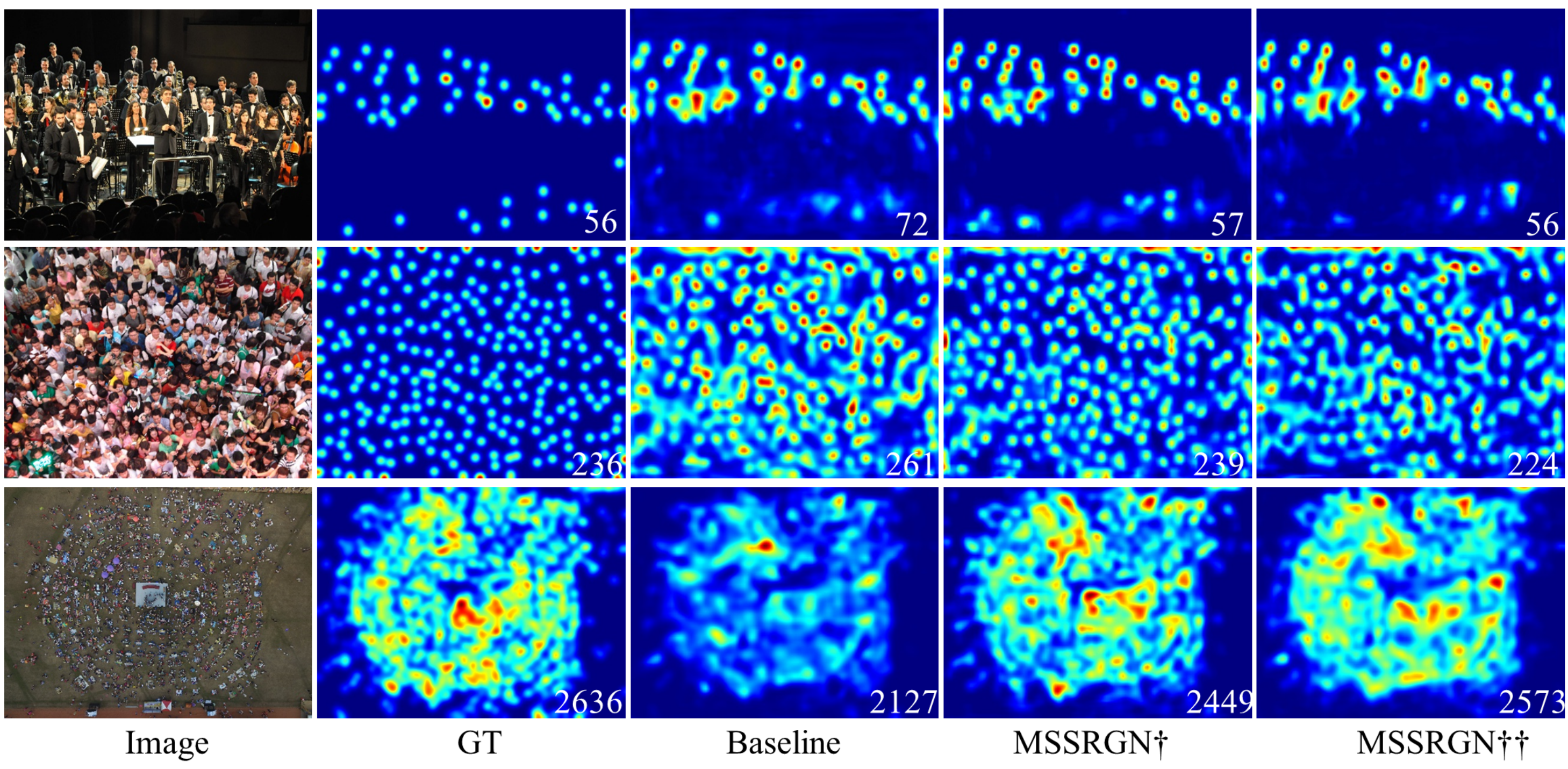}
    }
\vspace{-10pt}
\caption{
Visualizations on the SR-Crowd dataset. The MSSRGN presents sharper density maps than the baseline method.
}
\centering
\label{fig:visualization}
\vspace{-5pt}
\end{figure}



\section{Experiments}
This section introduces implementation details, evaluation metrics, ablation studies, and comparisons.

\textbf{Implementation Details.}
The MSSRGN is an end-to-end structure. We flip each image randomly. The first 10 convolutional layers are fine-tuned on the basis of the VGG-16 pre-training model. The other layers are initialized by a Gaussian initialization with a 0.01 standard deviation. We use the adaptive moment estimation optimizer. The learning rate as $10^{-5}$. The number of training epochs is set to 500. 

\textbf{Evaluation Metrics.}
We generally use the mean absolute error (MAE) and the root mean squared error (RMSE) as evaluation metrics in the crowd counting task. Their definition is as
$MAE=\frac{1}{M} \sum_{j=1}^{M}\left|b_{j}-b_{j}^{G T}\right|$,
$RMSE=\sqrt{\frac{1}{M} \sum_{j=1}^{M}\left|b_{j}-b_{j}^{G T}\right|^{2}}$,
where $M$ is the number of test images, $b_{j}^{G T}$ is the ground truth of counting, $b_{j}$ represents the estimated count in the $j$-th image.

\textbf{Ablation Studies.} We collect LR images from the original images using down-sampling methods. We first compare the performance of different down-sampling strategies. As shown in Table~\ref{tab:interpolation}, linear interpolation has the best performance. 
It achieves 5.959 and 2.224 lower MAE than the cubic and lanczos4 interpolation. 
We employ the linear interpolation in our experiments.
Feature fusion of different scales facilitates the extraction of multi-scale features. We then conduct experiments to analyze the effectiveness of different features fusion. As shown in Table~\ref{tab:stage}, the MAE of fusing stages 3, 4, 5 is 148.171. It achieves 3.499 lower MAE and 55.480 lower RMSE than the second place. We fuse stages 3, 4, and 5 in the following.
We present visualizations from the baseline method and the MSSRGN in Fig.~\ref{fig:visualization}. Three images are captured in the sparse, crowded, and dense senses, respectively. As we can see that the baseline method predicts density maps with serious noises in background regions. The proposed MSSRGN output density maps clearer and sharper.

\textbf{Comparisons.} 
As shown in Table~\ref{tab:comparisons}, we evaluate our method on the SR-Crowd, ShanghaiTech Part\_A, and UCF-QNRF datasets. The experimental results demonstrate that our MSSR module is plug-in plug-out and efficient. Specifically, on the SR-Crowd dataset, for CSRNET and FPN, MAE is reduced by 7.284 and 6.807, respectively, when adopting the upscale factor as 2$\times$. For the 4$\times$ upscale factor, MAE is reduced by 8.705 and 1.631, respectively. We also conducted comparative experiments with CLTR~\cite{liang2022end} and MAN~\cite{lin2022boosting} on the SR-Crowd, which achieve 170.684 and 151.890 MAE, respectively. For the ShanghaiTech Part\_A and the UCF-QNRF dataset, we downsample the images 2$\times$ and 4$\times$ as input. Use the original images as the ground truth for the SR task. Resize the original coordinates of head centers according to the upscale factors to generate new gt centers. On the ShanghaiTech Part\_A, the MSSRGN\dag~and MSSRGN\dag\dag~achieve 6.506 and 6.687 MAE improvement. 
We also conduct experiments on the UCF-QNRF dataset, a challenging dataset with dramatic variations both in crowd density and image resolution. Concretely, we resize the long side of images to 512, 1024, and 2048. And resize the short of images corresponding multiples to mitigate the enormous resolution variations. As shown in Table~\ref{tab:comparisons}, the MSSRGN\dag~achieves the best performance in the UCF-QNRF dataset. It reduces the MAE and RMSE values by 3.379 and 3.495. These experiments demonstrate the effectiveness of our method. 

\section{Conclusion}
In this paper, we propose to employ the SR task as an auxiliary task to deal with the LR circumstances in crowd counting. We propose an elegant method, MSSRM, to overcome the low-resolution barriers with no inference cost. The MSSRM is plug-in plug-out. It works in the training phase and guides the network to estimate the lost details as compensation for the blurry presentations. Besides, we propose an SR-Crowd dataset containing hierarchical-resolution images to achieve the SR task. Extensive experiments on SR-Crowd, ShanghaiTech Part\_A, and UCF-QNRF datasets demonstrate the effectiveness of our method. We believe our contributions broaden the application potentials of existing crowd counting methods. 


\clearpage



\vfill\pagebreak



\bibliographystyle{IEEEbib}
\bibliography{strings,refs}

\end{document}